\documentclass{article}

\usepackage{arxiv}

\usepackage[utf8]{inputenc} 
\usepackage[T1]{fontenc}    
\usepackage{hyperref}       
\usepackage{url}            
\usepackage{booktabs}       
\usepackage{amsfonts}       
\usepackage{nicefrac}       
\usepackage{microtype}      
\usepackage{lipsum}

\usepackage[bottom]{footmisc}

\usepackage{setspace} 
\usepackage[utf8]{inputenc} 
\usepackage{amsmath, amsthm, amssymb}	
\usepackage{amssymb}               
\usepackage{graphicx}              
\usepackage{epstopdf}
\usepackage{commath}
\usepackage{lscape}
\usepackage{amsfonts}
\usepackage{icomma} 
\usepackage[small,bf,hang]{caption}

\usepackage{algorithm}
\usepackage[noend]{algpseudocode}

\usepackage{booktabs}
\usepackage{pgfplots}
\usepackage{pgfplotstable}
\usepackage{longtable}
\usepackage{lscape}
\usepackage{subcaption}
\makeindex

\author{
	Tom Vermeire\thanks{Corresponding author.} \\
	Department of Engineering Management\\
	University of Antwerp\\
	\And
	David Martens \\
	Department of Engineering Management\\
	University of Antwerp\\
}

\begin{document}
\title{Explainable Image Classification with \\ Evidence Counterfactual}
\maketitle


\begin{abstract}
The complexity of state-of-the-art modeling techniques for image classification impedes the ability to explain model predictions in an interpretable way. 
Existing explanation methods generally create importance rankings in terms of pixels or pixel groups.
However, the resulting explanations lack an optimal size, do not consider feature dependence and are only related to one class. 
Counterfactual explanation methods are considered promising to explain complex model decisions, since they are associated with a high degree of human interpretability.
In this paper, SEDC is introduced as a model-agnostic instance-level explanation method for image classification to obtain visual counterfactual explanations.
For a given image, SEDC searches a small set of segments that, in case of removal, alters the classification.
As image classification tasks are typically multiclass problems, SEDC-T is proposed as an alternative method that allows specifying a target counterfactual class.
We compare SEDC(-T) with popular feature importance methods such as LRP, LIME and SHAP, and we describe how the mentioned importance ranking issues are addressed. 
Moreover, concrete examples and experiments illustrate the potential of our approach (1) to obtain trust and insight, and (2) to obtain input for model improvement by explaining misclassifications.

\end{abstract}

\section{Introduction} \label{sec:introduction}

The use of advanced machine learning (ML) techniques for image classification has known substantial progress over the past years.
Predictive performance has significantly improved, mainly due to the use of deep learning~\cite{lecun2015}.
However, this increased performance comes at a cost of increased model complexity and opacity.  
As a result, state-of-the-art image classification models are used in a black-box way, without being able to explain model decisions.

The need for explainability has become an important topic, often referred to as explainable artificial intelligence (XAI)~\cite{goebel2018, gunning2017, adadi2018, miller2018}.
Often cited motivations are increased trust in the model, compliance with regulations and laws, derivation of insights and guidance for model debugging~\cite{goebel2018, doshivelez2017}. 
A lack of explainability is considered a major barrier for the adoption of automated decision making by companies~\cite{samek2019, barocas2019, chander2018}. 
Recent governmental initiatives point at the importance of explainability~\cite{goodman2017, gunning2017}.
In the European Union, XAI has become an integral part of recent legal and ethical frameworks, such as the General Data Protection Regulation (GDPR)~\cite{goodman2017} and the guidelines on ethics in artificial intelligence~\cite{madiega2019}. 
We especially refer to the ``right to explanation", which is defined as the formal right to get an explanation for decisions that affect them in a legal or financial way.
It is argued that this right is part of GDPR~\cite{goodman2017}, although this view is much debated~\cite{wachter2017b}.   
Apart from regulatory pressure, companies face the increased awareness of consumers regarding privacy and automated decision-making. 

In general, an explanation provides an answer to the question why an event occurred. 
In classification tasks, the event corresponds to an instance being assigned to a specific class. 
Closely related to explainability is the concept of interpretability, which can be defined as the property of being understandable for humans.
The terms explainability and interpretability are often used interchangeably, although some authors differentiate between them~\cite{lipton2016, gilpin2018}. 
In line with Lipton~\cite{lipton2016}, we follow the reasoning that a good explanation leads to post-hoc interpretability.

As image classification for critical decisions is gaining ground, explainability is also becoming important in that context. 
We can think of applications such as medical image diagnosis~\cite{esteva2017} or damage assessment in insurance~\cite{shang2018}, for which patients or consumers obviously demand an explanation.
Explainability becomes even more important when severe misclassifications occur~\cite{simonite2018, lee2019}.
Besides physical and/or reputational damage caused by the misclassification itself, companies should be able to explain what went wrong, at least to prevent this from happening in the future and to restore trust.

For image classification, it can be argued that a good explanation allows to reveal an understandable pattern that led to the classification. 
If the pattern of a correct classification is true in the real world, this contributes to trust in the model and insight in the decision.
Furthermore, a good explanation should show for misclassifications why the error was made and provide input for model improvement. 
Third, an explanation can also reveal that a correct classification has occurred for wrong reasons, which cannot be derived from the black-box prediction itself.
Examples can be found where bias in the data was learned during model training~\cite{lapuschkin2019}.
Also here, the discovered pattern is insightful en can be used for model improvement. 

Multiple explanation methods for (image) classification have been proposed in the literature. 
A distinction can be made between global explanations, which apply to a model in general, and instance-level explanations, which focus on isolated model predictions~\cite{martens2016edc}. 
Furthermore, methods to generate explanations can be model-specific (i.e., specific for the type of model, also called decompositional) or model-agnostic (i.e., applicable to different types of models, also called pedagogical). 
Although model-specific explanation methods can be more accurate and efficient, the main advantage of model-agnostic methods is their flexibility~\cite{ribeiro2016}. 

Popular instance-level explanation methods for image classification such as LIME~\cite{ribeiro2016lime}, SHAP~\cite{lundberg2017} and LRP~\cite{bach2015}, typically create feature importance rankings. 
Although insightful, these methods have clear drawbacks: they do not determine the optimal explanation size, they do not account for feature dependence, and they are related to only one prediction class. 

In general, counterfactual explanations are considered promising to generate interpretable instance-level explanations of complex models~\cite{martens2016edc, wachter2017}.
Advantages are the contrastive nature~\cite{lipton1990}, possible compliance with regulations~\cite{wachter2017}, no constraints on model type and model complexity, and no need for model disclosure~\cite{barocas2019}.
Building on previous work, we apply counterfactual reasoning to explain individual image classifications.

The contributions of this paper are threefold. 
First, we introduce novel model-agnostic methods, SEDC and SEDC-T, to generate instance-level explanations for image classification. 
The visual and counterfactual nature of these explanations is strongly believed to increase interpretability.
Second, we argue how this method addresses limitations of existing explanation methods. 
Third, we show with concrete examples and experiments how this method is promising for both increasing trust and insight, and improving models.

\section{Related Work} \label{sec:related_work}

Over the past years, several techniques were introduced to explain individual predictions of complex image classification models.
This section gives an overview of earlier work that is relevant for the purpose of this paper. 

\subsection{Existing explanation methods} \label{subsec:existing_methods}

LIME~\cite{ribeiro2016lime} and the model-agnostic implementation of SHAP~\cite{lundberg2017} are two comparable methods that can be used to generate instance-level explanations for image classification.
For both methods, an image of interest is segmented, a number of perturbed samples are created by switching off segments, and these samples are predicted by the original classification model.
In LIME, the predictions of this artificial data set are then used as labels to train an interpretable linear model. 
The samples are weighted according to the proximity to the instance of relevance.
In SHAP, Shapley values of the segments are approximated, which reflect their contribution to the prediction.
The interpretable (linear) model around the prediction (LIME) or the Shapley values (SHAP) can be translated into a segment importance ranking.
Finally, a visual explanation can be obtained by showing the most (and least) important segments of the image.
The model-agnostic approach of these methods offers a high flexibility regarding data and model types.
However, an important issue for these approaches is that the resulting explanations are found to be unstable.
Small perturbations of an image with little or no impact on the model prediction, can lead to considerably different explanations~\cite{alvarezmelis2018}. 
Another source of instability is the random sampling process to generate image perturbations.  
This can lead to different explanations for a given image and a given model, which is clearly undesirable.

Another strategy to create instance-level explanations is calculating the importance of the individual input pixels for a certain class.
These can be translated into visual explanations by creating heat maps. 
One approach is the local perturbation of images and the direct measurement of the influence on prediction scores.
Zeiler and Fergus~\cite{zeiler2014} measure the impact of graying out squared segments in an image to identify the most important pixels supporting a certain class. 
A comparable method is used by Zintgraf et al.~\cite{zintgraf2017} who replace squared segments by using conditional sampling, instead of graying them out.
A second approach is taken by Bach et al.~\cite{bach2015} who introduce Layer-Wise Relevance Propagation (LRP) as a model-specific method to create instance-level explanations for neural networks. 
The importance of individual input pixels for a certain class is calculated by redistributing the prediction scores backwards through the network.
Compared to the first approach where every perturbation must be evaluated by the model, LRP is computationally more efficient because only one forward and one backward pass are needed for an image.
A disadvantage of these pixel-wise heat map methods is the low abstraction level of the explanations~\cite{samek2019}.
Since individual pixels are typically meaningless for humans, it is not always straightforward to derive interpretable concepts from it.

Common to the methods outlined above is that they generate explanations in the form of a feature importance for the prediction of interest. 
The features can be either individual pixels or pixel segments. 
As argued by Fernandez et al.~\cite{fernandez2020}, a disadvantage of such rankings is that they do not reveal what is minimally needed to alter a prediction. 
This relates to the fact that the size of an explanation (number of pixels or segments) is not optimized, since only a relative importance ordering is provided. 
A second drawback is that these rankings do not account for features being dependent on each other for the prediction~\cite{fernandez2020}. 
For instance, it is possible that a certain segment is only important due to the presence of another segment (that is possibly lower-ranked).
These dynamics cannot be derived solely from the ordered features.
As a third drawback we refer to the fact that importance rankings relate to one prediction class, which is insufficient when one wants to know why a class is chosen over another in case of multiclass classification.
They give an idea of evidence for or against this class, without making a direct link with other possible classes. 
Although it is possible to create rankings for different prediction classes, it is not always straightforward to delineate the discriminative features (which we illustrate in Section~\ref{sec:results}). 

\subsection{Counterfactual Reasoning for Image Classification} \label{subsec:counterfactual}

Many authors in the field of philosophy and cognitive science have raised the importance of contrastive explanations~\cite{miller2018}.
As argued by Lipton~\cite{lipton1990}, it is easier for humans to think in terms of contrastive explanations instead of giving a full causal attribution to an event.
This means that in case we want to know why a certain event took place, we actually wonder why that event occurred rather than another.
That other event is often referred to as the counterfactual case~\cite{lewis1974, mandel2005}.

The notion of counterfactual explanations is considered promising to explain decisions of complex machine learning models~\cite{martens2016edc, wachter2017}.
Such an explanation can be defined as a set of features that would alter the classification, if they were not present.
Martens and Provost~\cite{martens2016edc} were the first to apply this idea for predictive modeling, in the context of document classification.
Similar approaches were taken for classification based on fine-grained behavioral data~\cite{chen2017, ramon2019} and structured data~\cite{fernandez2020}. 
Apart from the contrastiveness, counterfactual explanations have other benefits. 
It is argued that they are more likely to comply with recent regulatory developments such as GDPR.
Wachter et al.~\cite{wachter2017} state that counterfactual explanations are well-suited to fill three important needs of data subjects: information on how a decision was reached, grounds to contest adverse decisions and an idea of what could be changed to receive a desired outcome.
Moreover, formulating an explanation as a set of features does not put constraints on model type and complexity~\cite{barocas2019}, which should make it robust for developments in modeling techniques.
Finally, the explanation can be done without disclosing the entire model~\cite{barocas2019}, which allows companies to give only the necessary information without revealing trade secrets. 

Several authors have used approaches that are closely related to counterfactual reasoning for image classification. 
Adversarial example methods aim at finding very small image perturbations that lead to false classifications~\cite{szegedy2013, goodfellow2014, su2019}.
This has proven useful to protect a classifier against attempts to deceive it.
However, since the found perturbations are often too small to be visible for humans (in extreme cases only one pixel), they cannot be used as an interpretable counterfactual explanation.

Dhurandhar et al.~\cite{dhurandhar2018} introduce a method for creating contrastive image classification explanations consisting of two parts: the pixels that should be minimally present for the original classification and the pixels that should be minimally absent for another classification.
Also in this case, the found explanations can be very small and/or hard to interpret. 
Moreover, to verify whether pixels should be absent, it is necessary to add them in some way. 
Dhurandhar et al.~\cite{dhurandhar2018} do this for the well-known MNIST dataset of hand-written digits. 
Since this dataset contains grayscale images, adding pixels is rather straightforward (i.e., adding white on a black background).
However, for colored images the idea of adding pixels is less obvious and not applied in practice.

To overcome the difficulties concerning image perturbation and small visual explanations, the use of natural language was proposed by Hendricks et al.~\cite{hendricks2018} to generate textual counterfactual explanations.
This method results in a textual description of evidence that, in case of presence, would change the image classification.
However, this relies on human description of the relevant concepts present in an image, which is considered difficult to obtain and objectify.
Goyal et al.~\cite{goyal2019} generate visual counterfactual explanations for a chosen counterfactual class by replacing parts of an image by parts of an image belonging to the counterfactual class.
Hence, this requires access to image data of that counterfactual class, which is not considered feasible in all contexts. 

Other authors specifically aim at identifying the parts of an image that should be minimally present for its classification.
Zhou et al.~\cite{zhou2014} iteratively gray out the least informative segments of a correctly classified image until it is misclassified.
Ribeiro et al.~\cite{ribeiro2018} proposed an algorithm to localize the segments of an image that are sufficient for its classification, irrespective of perturbations of the remaining segments.
Hence, this algorithm searches for evidence that rules out any counterfactual class.
Although both approaches use counterfactual reasoning to some extent, their goal is not to find a counterfactual explanation.

It can be concluded that current counterfactual methods have issues to generate interpretable instance-level explanations for image classification. 
Therefore, we propose a novel method that results in enhanced explanations, only based on the image of interest and without the need of human intervention.

\section{Search for EviDence Counterfactual for Image Classification} \label{sec:SEDC}

Martens and Provost~\cite{martens2016edc} introduce a model-agnostic search algorithm (SEDC) to find counterfactual explanations for document classifications.
An explanation can be seen as an irreducible set of features (i.e. words) that, in case they were not present, would alter the document classification.
In this context, irreducible means that removing any subset of the explanation would not change the classification.
We explore how an adapted version of this method can be used to generate visual counterfactual explanations for image classification. 
In the remainder of this paper we will refer to this algorithm as Search for EviDence Counterfactual (SEDC\footnote{We pronounce this as `Set See'.}).
 
Consider an image $I$ assigned to class $c$ by a classifier $C_M$.
In line with Martens and Provost~\cite{martens2016edc}, the objective is to find a counterfactual explanation $E$ of the following form: an irreducible set of segments that leads to another classification after removal. The segmentation and removal of segments will be discussed in Section~\ref{subsec:building_blocks}.

This definition can be formalized as:

\begin{equation}
E \subseteq I \; \textrm{(segments in image)}
\end{equation}
\begin{equation}
C_M(I \setminus E) \neq c \; \textrm{(class change)}
\end{equation}
\begin{equation}
\forall E' \subset E: C_M(I \setminus E') = c \; \textrm{(irreducible)}
\end{equation}

\begin{figure}
\centering
	\begin{subfigure}[t]{.3\linewidth}
		\centering
		\includegraphics[width=.9\linewidth]{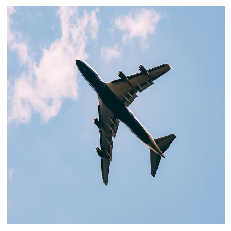}
		\caption{Predicted \\ class: \textit{warplane}}
		\label{subfig:warplane}
	\end{subfigure}
	\begin{subfigure}[t]{.3\linewidth}
		\centering
		\includegraphics[width=.9\linewidth]{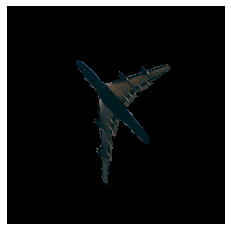}
		\caption{Counterfactual \\ explanation}
		\label{subfig:warplane_explanation}
		\end{subfigure}
	\begin{subfigure}[t]{.3\linewidth}
		\centering
		\includegraphics[width=.9\linewidth]{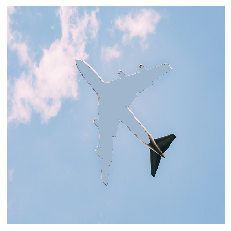}
		\caption{Counterfactual \\ class: \textit{wing}}
		\label{subfig:warplane_counterfactual}
		\end{subfigure}
\caption{Example of SEDC explanation.}		
\label{fig:warplane_sedc}
\end{figure}

The counterfactual explanation can be supplemented with the image causing the class change after removing the segments and the corresponding class.
An example of SEDC output is shown in Figure~\ref{fig:warplane_sedc}.
Figure~\ref{subfig:warplane} contains an image classified by a model as \textit{warplane}.
In Figure~\ref{subfig:warplane_explanation} the parts that lead to a class change after removal are shown.
Furthermore, Figure~\ref{subfig:warplane_counterfactual} gives the counterfactual case and class (\textit{wing} in this example). 

\subsection{SEDC for Image Classification} \label{subsec:SEDC}

In electronic format, an image is a collection of pixel values (one value per pixel for grayscale images, three values (RGB) per pixel for colored images).
These individual pixel values are typically used as input features for an image classifier. 
As interpretable concepts in images are embodied by groups of pixels or segments, we propose to perform a segmentation, similarly to LIME~\cite{ribeiro2016lime} and SHAP~\cite{lundberg2017}.
To this end, a label is assigned to each individual pixel reflecting the segment the pixel belongs to. 

In line with the reasoning behind SEDC for document classification, the goal is to find a small set of segments that would, in case of not being present, alter the image classification.
The original SEDC-algorithm~\cite{martens2016edc} was applied to binary document classifications tasks with a single prediction score reflecting the probability of belonging to the class of interest as output. 
In image classification applications, one is often confronted with more than two possible categories, each with its own prediction score.
Therefore, we generalize SEDC by enabling the occurrence of multiclass problems.
More specifically, additional segments are selected by looking for the highest reduction in predicted class score. 

A short version of the pseudo-code is outlined in Algorithm \ref{alg:SEDC_short}.
A more detailed version is outlined in Algorithm~\ref{alg:SEDC} (Appendix~\ref{app:sedc}).
SEDC takes an image of interest, an image classifier with corresponding scoring function and a segmentation as inputs, and produces a set of explanations as output.
Each individual explanation is a set of segments that leads to a class change after replacement.
A heuristic best-first search is performed in order to avoid a complete search through all possible segment combinations.
The best-first is each time selected based on the highest reduction in predicted class score and, subsequently, all expansions with one additional segment are considered. 
This search continues until one or more same-sized explanations are found after an expansion loop (i.e., the set of explanations is not empty).

In the procedure outlined above, already explored combinations remain part of the considered combinations to expand on. 
As a consequence, it is possible that, when searching for the best-first, the algorithm returns to a smaller combination in the search tree (for instance, after expanding combinations with three segments, the best-first might again be a combination of two segments). 
To assert that the algorithm does not get stuck in an endless loop by repeatedly returning to the same combination in the search tree, a selected combination is each time removed after all expansions with one additional segment are created.
We refer to this as the pruning step.
\begin{algorithm} 
	\caption{SEDC}
	\label{alg:SEDC_short}
	\begin{algorithmic}[0.6] \footnotesize
		\State \textbf{\underline{Inputs:}}
		\\
		$I$ \textcolor{gray}{\% Image to classify}
		\\
		${C}_{M}: I \rightarrow \{1,2,...,k\}$ \textcolor{gray}{\% Trained classifier with $k$ classes}
		\\
		$S = \{s_i, i = 1,2,...,l\}$ \textcolor{gray}{\% Segmentation of the image with $l$ segments} 
		\\
		\State \textbf{\underline{Procedure: }}
		\State $E = \{\}$ \textcolor{gray}{\% List of explanations}
		\State \textbf{for} $s_i$ in $S$ \textbf{do}
		\State\hspace{\algorithmicindent} \textbf{if} class change after removing $s_i$ from $I$ \textbf{then}
		\State\hspace{\algorithmicindent}\hspace{\algorithmicindent} $E = E \cup \{s_i\}$
		\State \hspace{\algorithmicindent} \textbf{end if}
		\State \textbf{end for}
		\State \textbf{while} $E = \emptyset$ \textbf{do}
		\State\hspace{\algorithmicindent} Select $best$ \textcolor{gray}{\% Best-first: segment set with highest reduction in predicted class score}
		\State\hspace{\algorithmicindent} $best\_set$ = expansions of $best$ with one segment
		\State\hspace{\algorithmicindent} \textbf{for} $C_0$ in $best\_set$ \textbf{do}
		\State\hspace{\algorithmicindent}\hspace{\algorithmicindent} \textbf{if} class change after removing $C_0$ from $I$ \textbf{then}
		\State\hspace{\algorithmicindent}\hspace{\algorithmicindent}\hspace{\algorithmicindent} $E = E \cup \{C_0\}$
		\State\hspace{\algorithmicindent}\hspace{\algorithmicindent} \textbf{end if}
		\State\hspace{\algorithmicindent} \textbf{end for}
		\State \textbf{end while}	
		\\
		\State \textbf{\underline{Output: }}
		\State Explanations in $E$
	\end{algorithmic}
\end{algorithm}

The final set of explanations can consist of one or more explanations, depending on how many expansions of the last combination to expand on result in a class change. 
In that case, we select the explanation $E$ with the highest reduction in predicted class score. 
Finally, an explanation can be visualized by only showing the corresponding segments of the image. 
Theoretically, it is possible that no class change occurs and, consequently, the while loop in the algorithm never ends.
To prevent this from happening, one or more additional conditions could be added to this while loop (e.g., maximum number of iterations, maximum computation time, etc.).

As shown in Martens and Provost~\cite{martens2016edc}, SEDC automatically results in irreducible explanations for linear classification models. 
This irreducibility cannot be guaranteed when using a nonlinear model, as is generally the case for image classification.  
Therefore, an additional local search can be performed by considering all possible subsets of the obtained explanation. 
If a subset leads to a class change after removal, the smallest set is taken as final explanation. 
When different subsets of equal size lead to a class change, the one with the highest reduction in predicted class score can be selected.

\subsection{SEDC with Target Counterfactual Class}

In SEDC, segments are iteratively selected based on the highest reduction in predicted class score until another classification is reached. 
We could also think of situations where it is useful to find counterfactual explanations for which the counterfactual class is predefined (not just any other class). 
For this purpose, we propose an alternative version SEDC-T\footnote{We pronounce this as `Set Seed'.} in which segments are iteratively removed until a predefined target class is reached.
A detailed version is outlined in Algorithm~\ref{alg:SEDC_target} (Appendix~\ref{app:sedc_target}).
The target class serves as an additional input parameter and segments are selected based on the largest difference between target class score and predicted class score. 
In case more than one explanation is found, the explanation leading to the highest increase in target class score can be selected.
Again, one or more additional conditions could be added to prevent the occurrence of an infinite while loop in case the target class is never reached.

SEDC-T allows for the generation of more nuanced explanations, since one can find out why the model predicts a class over another class of interest.
This can certainly be useful for explaining misclassifications.
In that case, it might be relevant to know why an image is not assigned to the correct class, rather than to know why it is assigned to the incorrect class.
Opposed to linear models for binary cases, these two notions are not the same in multiclass problems with nonlinear models.   
Therefore, by generating counterfactual explanations with the correct class as target counterfactual class, it is possible to identify those parts of the image that led to its misclassification. 

\subsection{Building blocks} \label{subsec:building_blocks}

The way the segmentation is created can take different forms.
For instance, one can choose for a (naive) squared segmentation by dividing the image in squares of equal size. 
However, the meaning of the resulting segments is highly dependent on the specific image.
Therefore, another option is using a more advanced segmentation algorithm that uses the numerical pixel values to obtain a more logical grouping in segments.

Furthermore, evidence being present or not should be further specified in the context of image data. 
Each time one or more segments are removed, an image perturbation is created. 
For textual and behavioral data, generating an instance wherein certain features are not present is usually straightforward (i.e., typically setting the corresponding values to zero). 
Setting the values of (groups of) pixels to zero in images corresponds to altering the color of the pixels to black (both for grayscale and colored images).
For images wherein black has a strong presence and/or meaning, this might be problematic since replacing the color by black has no or little impact.
The same applies for any other color chosen ex ante.
Alternatively, the segment replacement can be based on calculated pixel values. 
For instance, a segment can be changed to the mean/mode pixel values of the image as a whole, the segment itself or the neighboring segments.
Also, more advanced imputation methods for images are possible (e.g., blurring).
It can be expected that the replacement method is an important choice for the effectiveness of SEDC.
Blurring segments seems a reasonable choice when the presence or absence of details play an important role, while this is expected to be less effective in cases where shapes and edges are deciding.
In that case, averaging pixels out is considered more appropriate (e.g., \textit{warplane} example in Figure~\ref{fig:warplane_sedc}).

Hence, different possibilities exist for the segmentation method and the segment replacement method.
Both building blocks must be deliberately decided upon for an actual implementation and application of the algorithm.

\section{Results and Discussion} \label{sec:results}
In this section, the results of experiments with SEDC and SEDC-T for image classification are discussed.
First, we clarify the working of the algorithm with a detailed example on a \textit{chihuahua} image and we experimentally test the irreducibility of SEDC explanations. 
Second, SEDC(-T) explanations are compared with the feature importance ranking methods LRP, LIME and SHAP. 
Third, we show how our approach is promising for enhanced explainability: we illustrate how SEDC(-T) can lead to trust and insight in model decisions and we demonstrate the potential of SEDC-T for explaining misclassifications.
 
For these examples and experiments, SEDC and SEDC-T are implemented in \textit{Python}.
To compare with the feature importance ranking methods, we use the respective available implementations of LRP\footnote{\url{http://heatmapping.org}}, LIME\footnote{\url{https://github.com/marcotcr/lime}} and SHAP\footnote{\url{https://github.com/slundberg/shap}}. 
Google's pre-trained MobileNet V2 model is used as image classifier~\cite{sandler2018}. 
Since this neural network gives prediction scores for 1,001 different categories, the highest scoring class is selected as predicted class. 
Image data is downloaded from ImageNet~\cite{imagenet2020} and Kaggle~\cite{kaggle2014}. 
Our experiments are conducted on a laptop with Intel i7-8665U CPU (1.90 Ghz) and 16GB RAM.

\subsection{SEDC Applied in Practice} \label{subsec:sedc_example}

\subsubsection{Running Example}
\begin{figure}[!p]
\centering
	\begin{subfigure}[t]{.3\linewidth}
		\centering
		\includegraphics[width=.8\linewidth]{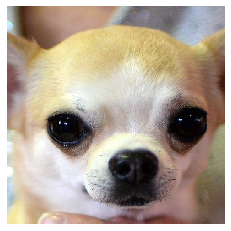}
		\caption{Predicted \\ class: \textit{chihuahua}}
		\label{subfig:chihuahua_original}
	\end{subfigure}
	\begin{subfigure}[t]{.3\linewidth}
		\centering
		\includegraphics[width=.8\linewidth]{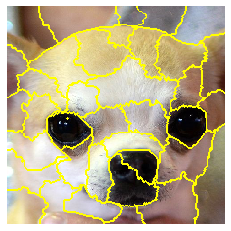}
		\caption{Segmentation}
		\label{subfig:chihuahua_segments}
	\end{subfigure}
		
	\begin{subfigure}[t]{.3\linewidth}
		\centering
		\includegraphics[width=.8\linewidth]{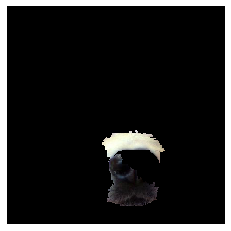}
		\caption{Counterfactual \\ explanation}
		\label{subfig:chihuahua_explanation}
	\end{subfigure}
	\begin{subfigure}[t]{.3\linewidth}
		\centering
		\includegraphics[width=.8\linewidth]{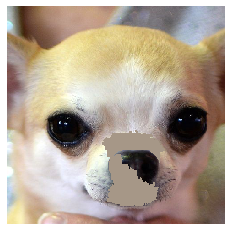}
		\caption{Counterfactual \\ class: \textit{French bulldog}}
		\label{subfig:chihuahua_counterfactual}
	\end{subfigure}
\caption{SEDC applied to \textit{chihuahua} image.}		
\label{fig:chihuahua}
\end{figure}

As a first illustration, we generate an explanation for an arbitrary image of a \textit{chihuahua} (see Figure~\ref{subfig:chihuahua_original}) wherein removing segments corresponds to replacing them by the mean pixel value of the image. 
Segments are created making use of the quick shift algorithm~\cite{vedaldi2008} (see Figure~\ref{subfig:chihuahua_segments}).
The resulting explanation is shown in Figure~\ref{subfig:chihuahua_explanation}.
Figure~\ref{subfig:chihuahua_counterfactual} contains the perturbed image leading to a class change.
After replacing two segments near the nose of the \textit{chihuahua}, the image is classified as \textit{French bulldog}. 

\begin{figure}[!p]
\centering	
	\begin{subfigure}[t]{.3\linewidth}
		\centering
		\includegraphics[width=.8\linewidth]{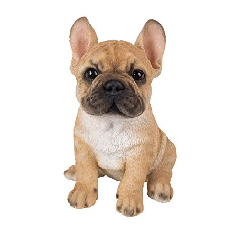}
		\caption{\textit{French bulldog}}
		\label{subfig:french_bulldog}
	\end{subfigure}
	\begin{subfigure}[t]{.3\linewidth}
		\centering
		\includegraphics[width=.8\linewidth]{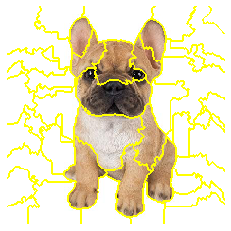}
		\caption{\textit{Segmentation}}
		\label{subfig:french_bulldog_segments}
	\end{subfigure}	
		
	\begin{subfigure}[t]{.3\linewidth}
		\centering
		\includegraphics[width=.8\linewidth]{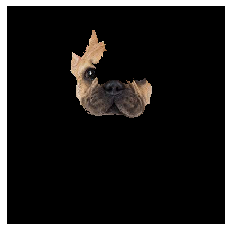}
		\caption{Counterfactual \\ explanation}
		\label{subfig:french_bulldog_sedc_target}
	\end{subfigure}
	\begin{subfigure}[t]{.3\linewidth}
		\centering
		\includegraphics[width=.8\linewidth]{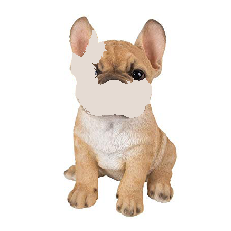}
		\caption{Counterfactual \\ class: \textit{chihuahua}}
		\label{subfig:french_bulldog_sedc_target_counterfactual}
	\end{subfigure}
\caption{SEDC-T applied to \textit{French bulldog} image.}		
\label{fig:french_bulldog}
\end{figure}

It is interesting to compare the obtained explanation and perturbed image with an image of an actual \textit{French bulldog} (see Figure~\ref{subfig:french_bulldog}).
At first glance, someone would probably identify the eyes of this \textit{chihuahua} as the most important segments. 
However, the eyes are not that different from the eyes of a \textit{French bulldog}. 
It can be argued that the nose is the most distinguishing characteristic between the two classes, as indicated by SEDC.

In addition, we apply SEDC-T to the \textit{chihuahua} example, this time in the opposite direction, to answer the question: ``Why is this image classified as a \textit{French bulldog}, and not as a \textit{chihuahua}?" 
The \textit{French bulldog} image is taken as the input image and \textit{chihuahua} is taken as target counterfactual class. 
This results in the explanation and perturbed image in Figure~\ref{subfig:french_bulldog_sedc_target} and Figure~\ref{subfig:french_bulldog_sedc_target_counterfactual}.
Also here, the nose of the dog is considered the most distinguishing characteristic, supporting the previous explanation. 

Alternative implementations with two other segmentation methods have also been applied (squared and SLIC~\cite{achanta2012}).
In both cases the resulting explanations point at the nose of the \textit{chihuahua} as well.
Finally, implementations with other segment replacement methods (random pixels and segment blurring by applying Gaussian smoothing~\cite{haddad1991}) consistently point to the nose of the \textit{chihuahua} as part of the counterfactual explanation.

\subsubsection{Irreducibility}

As mentioned in Section~\ref{subsec:SEDC}, SEDC cannot guarantee that the obtained explanation is irreducible when a nonlinear model is used. 
This can be verified by evaluating whether any subset of the explanation leads to a class change. 
We applied this to 100 images of \textit{cats}~\cite{kaggle2014}. 
A smaller explanation was found in two cases, while in 96 cases the explanation was found to be irreducible. 
In the two other cases, the local search lasted longer than 15 seconds because of a massive number of possible subsets. 
This concerned an explanation with 16 segments and one with 23 segments, implying that respectively 65,518 and 8,388,583 subsets must be considered to assess irreducibility. 
Since the irreducibility is almost never violated (only 2\% in our experiment) and the additional local search can be very time consuming for larger explanations, it can be argued that this step could be omitted or considered an optional post-processing step. 

\subsection{Benchmark against Feature Importance Ranking Methods}

\subsubsection{Comparison with Heat Maps}

\begin{figure}[H]
\centering
	\begin{subfigure}[t]{.3\linewidth}
		\centering
		\includegraphics[width=.8\linewidth]{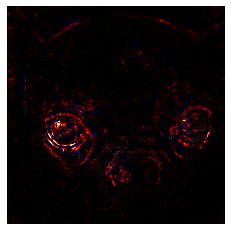}
		\caption{LRP heat map for \textit{chihuahua}}
		\label{subfig:heatmap_chihuahua}
	\end{subfigure}
	\begin{subfigure}[t]{.3\linewidth}
		\centering
		\includegraphics[width=.8\linewidth]{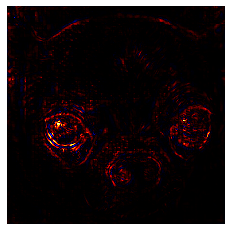}
		\caption{LRP heat map for \textit{French bulldog}}
		\label{subfig:heatmap_chihuahua_frenchbulldog}
	\end{subfigure}
	\begin{subfigure}[t]{.3\linewidth}
		\centering
		\includegraphics[width=.8\linewidth]{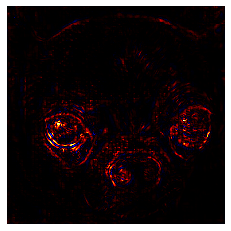}
		\caption{LRP heat map for \textit{tennis ball}}
		\label{subfig:heatmap_chihuahua_tennisball}
	\end{subfigure}
\caption{LRP applied to \textit{chihuahua} image.}		
\label{fig:lrp_explanations}
\end{figure}

LRP heat maps~\cite{bach2015} for the \textit{chihuahua} image are shown in Figure~\ref{fig:lrp_explanations}. 
They reveal some drawbacks compared to the counterfactual explanations.

First, the heat maps do not entail an explanation size. 
By coloring pixels according to the implied feature importance ranking, the user can get an idea of regions leading to the classification.
However, this type of explanation does not tell what is minimally needed.
The heat map does not allow to unambiguously assess whether the eyes are sufficient, or also the nose and contours of the head are needed.
In contrast, SEDC automatically limits the size of the explanation to the parts that would alter the classification.
This is considered useful, since the necessary size of an interpretable explanation can vary considerably between images.

Second, a heat map does not account for possible interdependence between features.
Figure \ref{subfig:heatmap_chihuahua} points to the importance of the eyes and the nose for the classification. 
Though, it does not tell whether the nose would also be important in case the eyes were not present. 
SEDC takes these possible dynamics into account by reevaluating the importance of the image segments after every removal.

Third, the heat map explanations only give evidence supporting one class and are thus not contrastive in nature.
Although it is possible to create heat maps for different classes and compare them, this is not always useful.
Consider for instance the heat maps of the \textit{chihuahua} image for the classes \textit{French bulldog} and \textit{tennis ball}, respectively shown in Figure~\ref{subfig:heatmap_chihuahua_frenchbulldog} and Figure~\ref{subfig:heatmap_chihuahua_tennisball}.
They can hardly be distinguished from the \textit{chihuahua} heat map and thus imply that the same image regions are important for all three classes (even for the \textit{tennis ball class}).
Therefore, it is impossible to derive why the image is classified as \textit{chihuahua} over \textit{French bulldog} or \textit{tennis ball}.
By contrast, SEDC bases its explanations on a class change and therefore searches for discriminative features (e.g., classified as \textit{chihuahua} over \textit{French bulldog} due to the nose).

\begin{figure}
\centering
	\begin{subfigure}[t]{.3\linewidth}
		\centering
		\includegraphics[width=.8\linewidth]{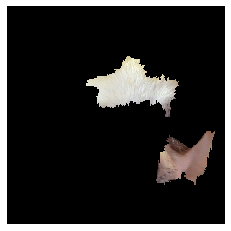}
		\caption{LIME explanation 1}
		\label{subfig:chihuahua_lime1}
	\end{subfigure}
	\begin{subfigure}[t]{.3\linewidth}
		\centering
		\includegraphics[width=.8\linewidth]{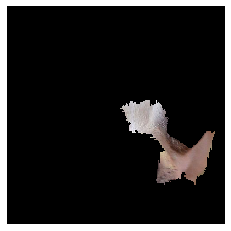}
		\caption{LIME explanation 2}
		\label{subfig:chihuahua_lime2}
		\end{subfigure}
	\begin{subfigure}[t]{.3\linewidth}
		\centering
		\includegraphics[width=.8\linewidth]{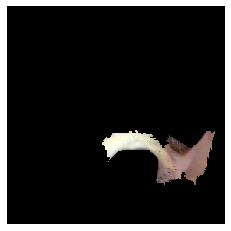}
		\caption{LIME explanation 3}
		\label{subfig:chihuahua_lime3}
	\end{subfigure}
	
	\begin{subfigure}[t]{.3\linewidth}
		\centering
		\includegraphics[width=.8\linewidth]{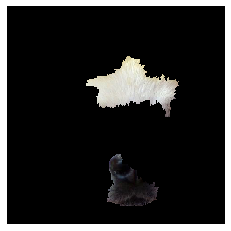}
		\caption{SHAP explanation 1}
		\label{subfig:chihuahua_shap1}
	\end{subfigure}
	\begin{subfigure}[t]{.3\linewidth}
		\centering
		\includegraphics[width=.8\linewidth]{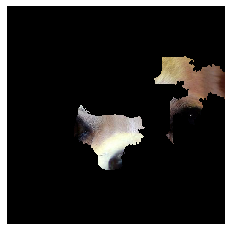}
		\caption{SHAP explanation 2}
		\label{subfig:chihuahua_shap2}
		\end{subfigure}
	\begin{subfigure}[t]{.3\linewidth}
		\centering
		\includegraphics[width=.8\linewidth]{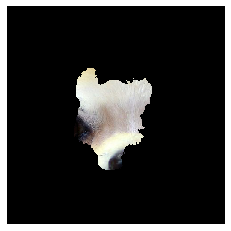}
		\caption{SHAP explanation 3}
		\label{subfig:chihuahua_shap3}
		\end{subfigure}
\caption{LIME and SHAP applied to \textit{chihuahua} image.}		
\label{fig:lime_shap_explanations}
\end{figure}

\subsubsection{Comparison with LIME and SHAP}

We also generate LIME and SHAP explanations for the \textit{chihuahua} image by making use of the quick shift segmentation.
A sample size of 1,000 image perturbations is used for both methods\footnote{This sample size is used in LIME and SHAP tutorials for image classification.}. 
Since applying SEDC results in an explanation consisting of two segments, only the two most important segments are shown for each method.
Remember that this is a parameter that a user needs to set for LIME and SHAP. 
Taking into account possible explanation instability due to the sampling process, the explanation generation process was repeated three times. 
The LIME and SHAP explanations are shown in Figure~\ref{fig:lime_shap_explanations}.

The fact that for both methods the three explanations differ for a fixed image and prediction model, illustrates the instability problem. 
A segment with a part of the \textit{chihuahua}'s cheek is part of each LIME explanation, while the second segment shows another characteristic of the dog. 
The first SHAP explanation points to the importance of the \textit{chihuahua}'s nose, the second contains a part of the eye and the third shifts the focus towards the forehead. 
In contrast, SEDC is deterministic and always results in the same explanation. 

The mentioned limitations of feature importance methods do also apply to LIME and SHAP: they do not provide an optimized explanation size (we took the size of SEDC explanations), the relative ordering of segments does not account for dependence between them, and the explanations are related to one class.
Regarding the latter, we revisit the idea of comparing explanations for different classes. 
First, it is not guaranteed that the explanations are sufficiently different to reveal discriminative regions.
Moreover, the instability issue adds a layer of complexity, since the explanation for each of the classes is subject to chance. 
This implies that one should compare multiple versions of explanations for each of the classes, which obviously renders the process more difficult and uncertain.

Since SEDC, LIME and SHAP generate visual explanations in a similar format (i.e., a set of segments), the resulting explanations can be further compared.
We conduct an experiment wherein SEDC, LIME and SHAP explanations are generated for 100 random images of \textit{cats}~\cite{kaggle2014}.
For the LIME and SHAP explanations, the same number of segments as in the corresponding SEDC explanation is taken.  
For each image 10 explanations per method (30 in total) are generated and, for each method, the following information is collected over 10 explanations: 

\begin{itemize}
	\item the stability in terms of Jaccard similarity~\cite{fletcher2018} (calculated as segments appearing in each of the 10 explanations divided by all unique segments in the 10 explanations),
	\item the average computation time,
	\item the counterfactual nature measured as fraction of explanations leading to a class change. 
\end{itemize}

Afterwards, the information on the image explanations is averaged over the 100 images.
The results are shown in Table~\ref{tab:experiment_sedc_lime_shap}.

\begin{table}[H]
\centering
\caption{Benchmarking results of SEDC, LIME \& SHAP explanations: stability (\%), mean and standard deviation of computation times (s), and counterfactual nature (\%).}
\label{tab:experiment_sedc_lime_shap}
\begin{tabular}{ccccc}
\textbf{Criterion} 		&						& \textbf{SEDC} & \textbf{LIME} & \textbf{SHAP} \\
\hline
stability (\%) 				&						& \textbf{100}	& 68.25					& 59.99 \\
\hline
computation time (s)	&						&               &               & \\
											& $\mu$			& \textbf{2.79}	& 26.13					& 34.75 \\
											& $\sigma$	& 4.64 					& \textbf{0.92} & 1.83 \\
\hline
counterfactual (\%)  	&						& \textbf{100}	& 48.70					& 61.90
\end{tabular}
\end{table}

First, the lower similarity of the generated LIME and SHAP explanations for a given image classification (respectively 68.25\% and 59.99\%) points to the instability issue. 
In contrast, the SEDC explanations for an individual image are always identical, since the approach is deterministic. 

Second, in terms of computation time, SEDC is generally faster than LIME and SHAP.
On average, SEDC finds an explanation in 2.79 seconds, while LIME and SHAP respectively need 26.13 and 34.76 seconds.
Only in extreme cases where a high number of segments must be removed to result in a class change, does SEDC take more time. 
This typically involves images wherein evidence supporting the predicted class is abundantly present and scattered across the image.
Although it is possible to speed up the computation time of LIME and SHAP by reducing the number of perturbed samples, this will lower the stability of the resulting explanations even further.
We note that the computation time of SEDC fluctuates more compared to LIME and SHAP.
Since the number of perturbations is chosen in advance for these methods, the time to compute explanations for different images is similar.
In contrast, SEDC generates additional perturbations until explanations are found.
For instance, since the explanation shown in Figure~\ref{subfig:chihuahua_explanation} consists of two segments on a total of 37, at least 73 perturbations are made and classified (37 with one perturbed segment and 36 with two perturbed segments).
This number could be higher in case other combinations with two segments were evaluated before returning to another one segment-combination in the search tree.  
Assuming SEDC explanations for this image would contain five segments, at least 175 perturbations would be needed to obtain them. 
As a result, the computation time for SEDC is generally more volatile than for LIME and SHAP.

Third, the most important segments resulting from LIME and SHAP only result in a class change after perturbation in respectively 48.70\% and 61.90\% of the cases. 
Remember this is always the case for SEDC, as it is the objective of the algorithm. 
This implies that, although LIME and SHAP identify segments that support the predicted class, these are not necessarily the most discriminative ones.
In other words, these explanations are not necessarily contrastive.

\subsection{SEDC(-T) for Enhanced Explainability}

\subsubsection{Trust and Insight}

The \textit{warplane} and \textit{chihuahua} examples illustrate how SEDC can contribute to the explainability objectives trust and insight. 
By identifying the discriminative segments of the image, SEDC shows that the model correctly classifies the warplane for the right reasons (trust) and reveals what differentiates the \textit{chihuahua} from a \textit{French bulldog} for the model (insight). 

\begin{figure}[!ph]
\centering
	\begin{subfigure}[t]{.3\linewidth}
		\centering
		\includegraphics[width=.8\linewidth]{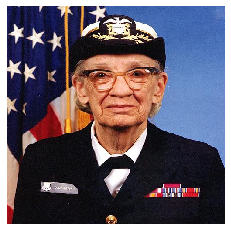}
		\caption{Predicted \\ class: \textit{military uniform}}
		\label{subfig:military_uniform_original}
	\end{subfigure}
	
	\begin{subfigure}[t]{.3\linewidth}
		\centering
		\includegraphics[width=.8\linewidth]{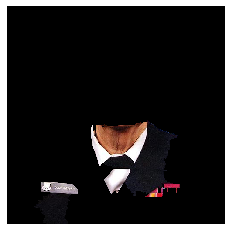}
		\caption{Counterfactual \\ explanation}
		\label{subfig:military_uniform_sedc}
		\end{subfigure}
	\begin{subfigure}[t]{.3\linewidth}
		\centering
		\includegraphics[width=.8\linewidth]{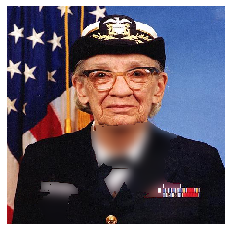}
		\caption{Counterfactual \\ class: \textit{mask}}
		\label{subfig:military_uniform_sedc_counterfactual}
	\end{subfigure}	
\caption{SEDC applied to \textit{military uniform} image.}		
\label{fig:military_uniform}
\end{figure}

As a next example, we consider an image that is classified as \textit{military uniform} shown in Figure~\ref{subfig:military_uniform_original}. 
Applying SEDC with blurred segment replacement results in the explanation and perturbed image shown in Figure~\ref{subfig:military_uniform_sedc} and Figure~\ref{subfig:military_uniform_sedc_counterfactual}.
The explanation points to the importance of the name tag, the medal ribbons, the bow tie and the neck of the person for the image being classified as \textit{military uniform}.
After blurring these parts, the image is classified as \textit{mask}.

\begin{figure}[!p]
\centering
	\begin{subfigure}[t]{.3\linewidth}
		\centering
		\includegraphics[width=.8\linewidth]{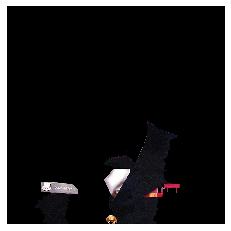}
		\caption{Counterfactual \\ explanation}
		\label{subfig:military_uniform_sedc_target_suit}
		\end{subfigure}
	\begin{subfigure}[t]{.3\linewidth}
		\centering
		\includegraphics[width=.8\linewidth]{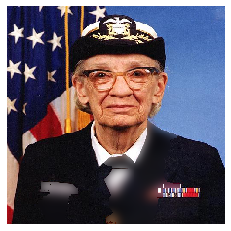}
		\caption{Counterfactual \\ class: \textit{suit}}
		\label{subfig:military_uniform_sedc_target_suit_counterfactual}
	\end{subfigure}
	
	\begin{subfigure}[t]{.3\linewidth}
		\centering
		\includegraphics[width=.8\linewidth]{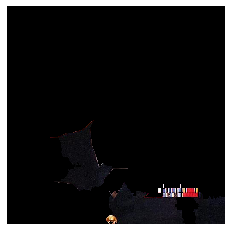}
		\caption{Counterfactual \\ explanation}
		\label{subfig:military_uniform_sedc_target_bowtie}
	\end{subfigure}
	\begin{subfigure}[t]{.3\linewidth}
		\centering
		\includegraphics[width=.8\linewidth]{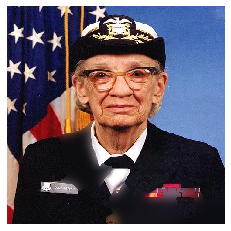}
		\caption{Counterfactual \\ class: \textit{bow tie}}
		\label{subfig:military_uniform_sedc_target_bowtie_counterfactual}
	\end{subfigure}
	
\caption{SEDC-T applied to \textit{military uniform} image.}		
\label{fig:military_uniform_target}
\end{figure}

One could also wonder what is needed to classify the image as \textit{suit} instead of \textit{military uniform} and apply SEDC-T with \textit{suit} as target counterfactual class.  
This results in the explanation and perturbed image shown in Figure~\ref{subfig:military_uniform_sedc_target_suit} and Figure~\ref{subfig:military_uniform_sedc_target_suit_counterfactual}.
In this case, the counterfactual explanation also points to the importance of the name tag, the medal ribbons and the buttons for classifying the image as \textit{military uniform} over \textit{bow tie}. In this case the neck is not part of the explanation.

We also apply SEDC-T with \textit{bow tie} as target and obtain the explanation and counterfactual shown in Figure~\ref{subfig:military_uniform_sedc_target_bowtie} and Figure~\ref{subfig:military_uniform_sedc_target_bowtie_counterfactual}. Here, the medal ribbons and the button are removed, but the bow tie is kept in place.
These examples illustrate again that SEDC and SEDC-T are useful to verify whether model predictions are made for the right reasons (trust).
The explanations contain segments with (interpretable) elements that are, according to the model, distinctive for a \textit{military uniform} over several other classes. 
This allows to assess the quality of the model decision, to reveal its decision boundaries and to derive insights from the data.
As a consequence, the (future) decisions of the model can be better understood.

\newpage
\subsubsection{Misclassifications and Model Improvement}

In Section~\ref{sec:introduction}, we argued that model improvement is an important explainability objective.
If it is possible to explain model errors (misclassifications), the explanation(s) can provide input to better understand why the model failed.
This information could then be used for model debugging (e.g., by gathering additional training data focusing on the identified relevant parts).
The next experiments demonstrate how SEDC-T can provide insights in such misclassifications.

\begin{figure}[!p]
\centering
	\begin{subfigure}[t]{.3\linewidth}
	\centering
	\includegraphics[width=.8\linewidth]{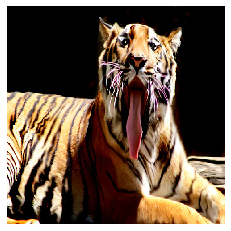}
	\caption{Predicted \\ class: \textit{tiger cat}}
	\label{subfig:tiger_target}
	\end{subfigure}
	\begin{subfigure}[t]{.3\linewidth}
	\centering
	\includegraphics[width=.8\linewidth]{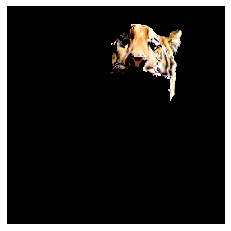}
	\caption{Counterfactual \\ explanation}
	\label{subfig:tiger_target_explanation}
	\end{subfigure}
	\begin{subfigure}[t]{.3\linewidth}
	\centering
	\includegraphics[width=.8\linewidth]{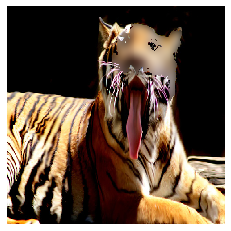}
	\caption{Counterfactual \\ class: \textit{tiger}}
	\label{subfig:tiger_target_counterfactual}
	\end{subfigure}
	
	\begin{subfigure}[t]{.3\linewidth}
	\centering
	\includegraphics[width=.8\linewidth]{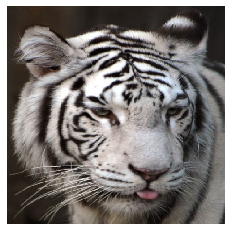}
	\caption{Predicted \\ class: \textit{tiger cat}}
	\label{subfig:tiger_target}
	\end{subfigure}
	\begin{subfigure}[t]{.3\linewidth}
	\centering
	\includegraphics[width=.8\linewidth]{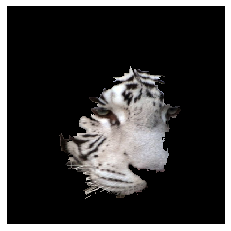}
	\caption{Counterfactual \\ explanation}
	\label{subfig:tiger_target_explanation}
	\end{subfigure}
	\begin{subfigure}[t]{.3\linewidth}
	\centering
	\includegraphics[width=.8\linewidth]{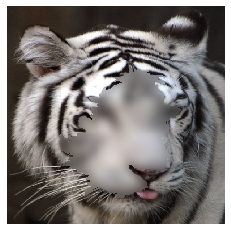}
	\caption{Counterfactual \\ class: \textit{tiger}}
	\label{subfig:tiger_target_counterfactual}
	\end{subfigure}
\caption{SEDC-T applied to \textit{tiger} classified as \textit{tiger cat}.}
\label{fig:tiger}
\end{figure}

We applied SEDC-T to 30 images of \textit{tigers} that where incorrectly classified as \textit{tiger cats}~\cite{imagenet2020}.
For 29 of these misclassified images, SEDC-T with \textit{tiger} as target class finds an explanation. 
Two examples are shown in Figure~\ref{fig:tiger}.
In these cases SEDC-T identifies (a) part(s) of the \textit{tiger} in the image as most important and discriminating (and not irrelevant parts such as the background). 
The model apparently has difficulties with discerning specific parts of the head and the fur pattern compared to those of a \textit{tiger cat}. 

\begin{figure}[!p]
\centering
	\begin{subfigure}[t]{.3\linewidth}
	\centering
	\includegraphics[width=.8\linewidth]{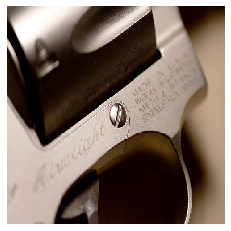}
	\caption{Predicted \\ class: \textit{pencil sharpener}}
	\label{subfig:revolver_target}
	\end{subfigure}
	\begin{subfigure}[t]{.3\linewidth}
	\centering
	\includegraphics[width=.8\linewidth]{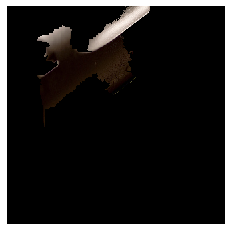}
	\caption{Counterfactual \\ explanation}
	\label{subfig:revolver_target_explanation}
	\end{subfigure}
	\begin{subfigure}[t]{.3\linewidth}
	\centering
	\includegraphics[width=.8\linewidth]{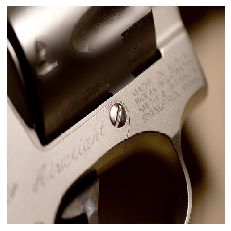}
	\caption{Counterfactual \\ class: \textit{revolver}}
	\label{subfig:revolver_target_counterfactual}
	\end{subfigure}
	
	\begin{subfigure}[t]{.3\linewidth}
	\centering
	\includegraphics[width=.8\linewidth]{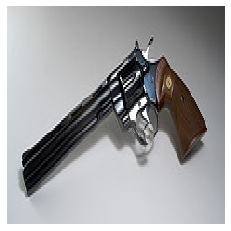}
	\caption{Predicted \\ class: \textit{pencil sharpener}}
	\label{subfig:revolver_target}
	\end{subfigure}
	\begin{subfigure}[t]{.3\linewidth}
	\centering
	\includegraphics[width=.8\linewidth]{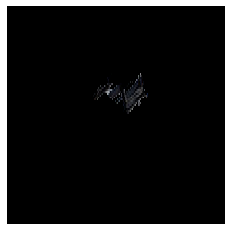}
	\caption{Counterfactual \\ explanation}
	\label{subfig:revolver_target_explanation}
	\end{subfigure}
	\begin{subfigure}[t]{.3\linewidth}
	\centering
	\includegraphics[width=.8\linewidth]{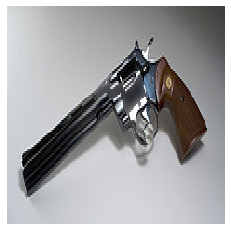}
	\caption{Counterfactual \\ class: \textit{revolver}}
	\label{subfig:revolver_target_counterfactual}
	\end{subfigure}
\caption{SEDC-T applied to \textit{revolver} classified as \textit{pencil sharpener}.}
\label{fig:revolver}
\end{figure}

\begin{figure}[!p]	
\centering
	\begin{subfigure}[t]{.3\linewidth}
	\centering
	\includegraphics[width=.8\linewidth]{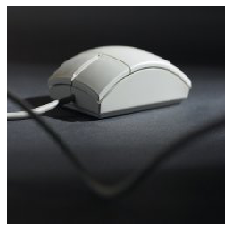}
	\caption{Predicted \\ class: \textit{soccer ball}}
	\label{subfig:mouse_target}
	\end{subfigure}
	\begin{subfigure}[t]{.3\linewidth}
	\centering
	\includegraphics[width=.8\linewidth]{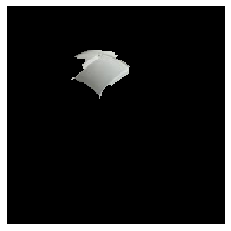}
	\caption{Counterfactual \\ explanation}
	\label{subfig:mouse_target_explanation}
	\end{subfigure}
	\begin{subfigure}[t]{.3\linewidth}
	\centering
	\includegraphics[width=.8\linewidth]{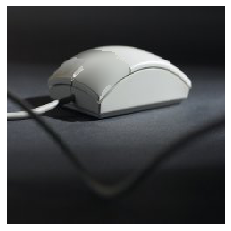}
	\caption{Counterfactual \\ class: \textit{mouse}}
	\label{subfig:mouse_target_counterfactual}
	\end{subfigure}
	
	\begin{subfigure}[t]{.3\linewidth}
	\centering
	\includegraphics[width=.8\linewidth]{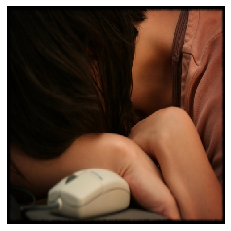}
	\caption{Predicted \\ class: \textit{soccer ball}}
	\label{subfig:mouse_target}
	\end{subfigure}
	\begin{subfigure}[t]{.3\linewidth}
	\centering
	\includegraphics[width=.8\linewidth]{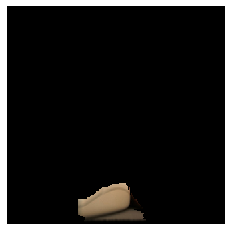}
	\caption{Counterfactual \\ explanation}
	\label{subfig:mouse_target_explanation}
	\end{subfigure}
	\begin{subfigure}[t]{.3\linewidth}
	\centering
	\includegraphics[width=.8\linewidth]{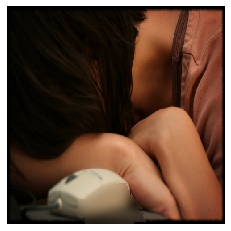}
	\caption{Counterfactual \\ class: \textit{mouse}}
	\label{subfig:mouse_target_counterfactual}
	\end{subfigure}
\caption{SEDC-T applied to \textit{mouse} classified as \textit{soccer ball}.}
\label{fig:mouse}
\end{figure}

\begin{figure}[!p]
\centering
	\begin{subfigure}[t]{.3\linewidth}
	\centering
	\includegraphics[width=.8\linewidth]{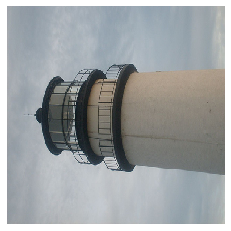}
	\caption{Predicted \\ class: \textit{missile}}
	\label{subfig:beacon_target}
	\end{subfigure}
	\begin{subfigure}[t]{.3\linewidth}
	\centering
	\includegraphics[width=.8\linewidth]{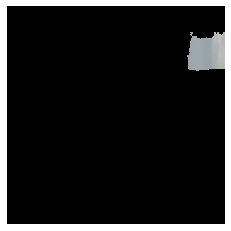}
	\caption{Counterfactual \\ explanation}
	\label{subfig:beacon_target_explanation}
	\end{subfigure}
	\begin{subfigure}[t]{.3\linewidth}
	\centering
	\includegraphics[width=.8\linewidth]{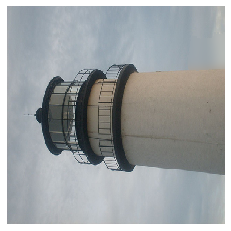}
	\caption{Counterfactual \\ class: \textit{beacon}}
	\label{subfig:beacon_target_counterfactual}
	\end{subfigure}
	
	\begin{subfigure}[t]{.3\linewidth}
	\centering
	\includegraphics[width=.8\linewidth]{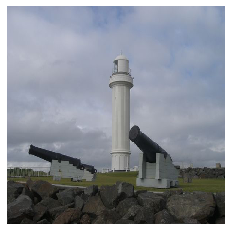}
	\caption{Predicted \\ class: \textit{missile}}
	\label{subfig:beacon_target}
	\end{subfigure}
	\begin{subfigure}[t]{.3\linewidth}
	\centering
	\includegraphics[width=.8\linewidth]{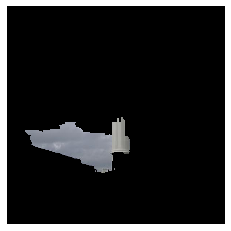}
	\caption{Counterfactual \\ explanation}
	\label{subfig:beacon_target_explanation}
	\end{subfigure}
	\begin{subfigure}[t]{.3\linewidth}
	\centering
	\includegraphics[width=.8\linewidth]{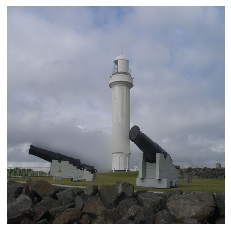}
	\caption{Counterfactual \\ class: \textit{beacon}}
	\label{subfig:beacon_target_counterfactual}
	\end{subfigure}
\caption{SEDC-T applied to \textit{beacon} classified as \textit{missile}.}
\label{fig:beacon}
\end{figure}

As a next example, SEDC-T is applied to images of a \textit{revolver} misclassified as \textit{pencil sharpener} (see Figure~\ref{fig:revolver}).
The counterfactual explanation identifies the cylinder of the \textit{revolver} as the reason for being mistaken as a \textit{pencil sharpener}.
We also applied SEDC-T to images of a computer \textit{mouse} misclassified as \textit{soccer ball} (see Figure~\ref{fig:mouse}).
In this case, the counterfactual explanation points to a part of the \textit{mouse}, which is presumably confused with the bounded faces on a \textit{soccer ball}. 
Another interesting example is the images of a \textit{beacon} misclassified as \textit{missile} shown in Figure~\ref{fig:beacon}.
The counterfactual explanation points to a part of the background with clouds resembling the exhaust plume of a \textit{missile}.

In order to more generally assess the feasibility of using SEDC-T to explain image misclassifications, we conduct an experiment on a larger scale.
Data sets with 20 different labels were downloaded from ImageNet~\cite{imagenet2020} (29,275 images in total).
We verified which images are misclassified by our model and selected the images belonging to the top five most occurring misclassifications per label. 
To these images (2,121), SEDC-T is applied with the correct class as target and a maximum search time per image of 15 seconds.
The results are summarized in Table~\ref{tab:misclassifications_short}.
In 86\% of the cases, SEDC-T finds an explanation leading to the correct class change. 
For the remaining 290 images, SEDC-T did not reach the correct class within the time limit.
A detailed overview of the data and the results per class is given in Table~\ref{tab:misclassifications} in \ref{app:data_misclassifications}.
\begin{table}[H]
\centering
\caption{Experiment: SEDC-T applied to misclassifications.}
\label{tab:misclassifications_short}
\begin{tabular}{ccc}
\textbf{\# images} 	& \textbf{Target found} & \textbf{Target not found} \\
\hline
2,121								& 1,831 (86\%)					& 290 (14\%)
\end{tabular}
\end{table}

In 289 of the 290 cases for which the target was not reached, SEDC-T does however result in a perturbation with an improved difference between the target and predicted class score after 15 seconds of search time.
Even though the correct class change is not (yet) reached, it is thus almost always feasible to find perturbations that lie closer to the correct class.

\begin{figure}[t]
\centering
	\begin{subfigure}[t]{.3\linewidth}
		\centering
		\includegraphics[width=.8\linewidth]{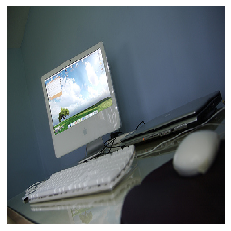}
		\caption{Predicted \\ class: \textit{desktop computer}}
		\label{subfig:mouse_desktop180}
	\end{subfigure}
	\begin{subfigure}[t]{.3\linewidth}
		\centering
		\includegraphics[width=.8\linewidth]{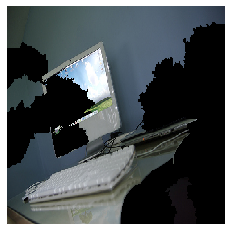}
		\caption{Counterfactual \\ explanation}
		\label{subfig:warplane_explanation}
		\end{subfigure}
	\begin{subfigure}[t]{.3\linewidth}
		\centering
		\includegraphics[width=.8\linewidth]{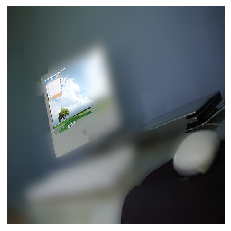}
		\caption{Counterfactual \\ class: \textit{mouse}}
		\label{subfig:mouse_desktop_perturbation180}
		\end{subfigure}
\caption{Misclassification for which SEDC-T without time limit succeeds.}		
\label{fig:mouse_desktop38}
\end{figure}

\begin{figure}[t]
\centering
	\begin{subfigure}[t]{.3\linewidth}
		\centering
		\includegraphics[width=.8\linewidth]{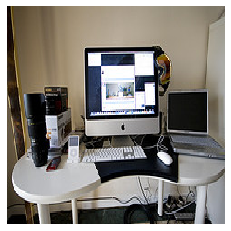}
		\caption{Predicted \\ class: \textit{desktop computer}}
		\label{subfig:mouse_desktop180}
	\end{subfigure}
	\begin{subfigure}[t]{.3\linewidth}
		\centering
		\includegraphics[width=.8\linewidth]{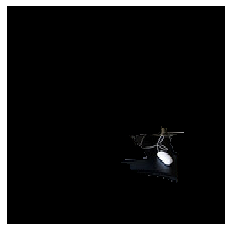}
		\caption{Counterfactual \\ explanation}
		\label{subfig:warplane_explanation}
		\end{subfigure}
	\begin{subfigure}[t]{.3\linewidth}
		\centering
		\includegraphics[width=.8\linewidth]{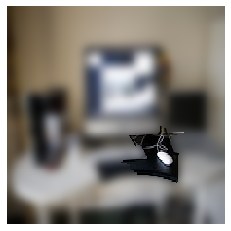}
		\caption{Counterfactual \\ class: \textit{fly}}
		\label{subfig:mouse_desktop_perturbation180}
		\end{subfigure}
\caption{Misclassification for which SEDC-T without time limit fails.}		
\label{fig:mouse_fly}
\end{figure}

Let us have a look at why the method fails to find an explanation.
In Table~\ref{tab:misclassifications_failed}, the misclassifications are shown for which SEDC-T is successful in less than 70\% of the cases and with an occurrence of at least 10.
\begin{table}[H]
\centering
\caption{Main cases where SEDC-T failed for misclassifications within time limit.}
\label{tab:misclassifications_failed}
\begin{tabular}{cccc}
\textbf{Class} 		& \textbf{Correct class} 	& \textbf{\# misclassified} & \textbf{\# target not found} \\
\hline
desk 							& mouse										& 57												& 36 (63\%) \\
desktop computer	& mouse										& 65												& 27 (42\%) \\
bearskin					& military uniform				& 33												& 13 (39\%)\\
\end{tabular}
\end{table}
For these cases, the labeling of the images seems to be the reason, since the supposedly wrong label could also be considered correct.
Images of \textit{mouses} classified as \textit{desktop computer} or \textit{desk} for which SEDC-T fails, are always ones wherein both objects are present.
Since in these cases, the \textit{mouse} is only a small element in the image, a lot of evidence must be removed to change the \textit{desktop computer} or \textit{desk} class. 
Likewise, a \textit{bearskin} is typically part of a \textit{military uniform}. 
In some cases, SEDC-T does find a solution when a longer search time is allowed. 
Consider for instance Figure~\ref{fig:mouse_desktop38}, where the target class is reached after 38 seconds. 
In other cases, the target class is never reached, even without a time limit.
Consider for instance Figure~\ref{fig:mouse_fly} where, even after removing all segments except for those containing the actual mouse, the image is still not classified as a \textit{mouse}.

The examples and the experiment show the potential of SEDC-T for explaining image misclassifications. 
In those situations, only evidence supporting the wrong class will not be sufficient as explanation, because one wants to know why the incorrect class was chosen over the correct class.
Existing explanation methods do not explicitly account for this, since they do not simultaneously consider multiple classes in the explanation process. 
In contrast, SEDC-T makes it possible to obtain the discriminative information, which can then be used to improve the existing classification model (e.g., providing it with more images of \textit{revolvers} focusing on the cylinder).

\section{Conclusions and Future Research}

Explaining predictions of complex image classification models is an important and challenging problem in the machine learning field. 
Current methods typically generate feature importance rankings, which have drawbacks regarding explanation size, feature dependence and being related to one prediction class.
In general, the notion of counterfactual explanations is considered promising to explain complex model decisions.
In this paper, SEDC and SEDC-T are proposed as model-agnostic instance-level explanation methods for image classification. 
They allow for the automatic generation of visual counterfactual explanations and address the mentioned issues of existing explanation methods.
Based on concrete examples, we compare SEDC with state-of-the-art techniques such as LIME, SHAP and LRP.
Moreover, we illustrate how our method contributes to three important explainability objectives: trust, insight and model improvement. 

Future research could focus on how a smarter search strategy can reduce computation times. 
For instance, it might be interesting to test how incorporating the segment or pixel importance rankings from existing explanation methods into the search for evidence counterfactual, could speed up the process. 
Another avenue of research is to further explore the segmentation and segment replacement methods and test their effectiveness in different application contexts. 
Finally, user studies with human subjects are needed to qualitatively test the interpretability and explainability of our approach.

\section{Acknowledgements}
This work was supported by the AXA Joint Research Initiative (JRI) project `Explainable Artificial Intelligence'.

\bibliographystyle{unsrt}
\bibliography{AISI-references}

\appendix

\section{SEDC algorithm} \label{app:sedc}
\begin{algorithm}[H]

	\caption{SEDC algorithm}
	\label{alg:SEDC}
	\begin{algorithmic}[0.6] \scriptsize
		\State \textbf{\underline{Inputs:}}
		\\
		$I$ \textcolor{gray}{\% Image to classify}
		\\
		${C}_{M}: I \rightarrow \{1,2,...,k\}$ \textcolor{gray}{\% Trained classifier with $k$ classes and scoring function $f_{C_M}$}
		\\
		$S = \{s_i, i = 1,2,...,l\}$ \textcolor{gray}{\% Segmentation of image with $l$ segments} 
		\\
		\State \textbf{\underline{Procedure:}}
		\State $c = {C}_{M}(I)$ \textcolor{gray}{\% Predicted class}
		\State $p_c = f_{C_{M,c}}(I)$ \textcolor{gray}{\% Score for predicted class}
		\State $E = \{\}$ \textcolor{gray}{\% List of explanations }
		\State $C = \{\}$ \textcolor{gray}{\% List of combinations to expand on }
		\State $P = \{\}$ \textcolor{gray}{\% List of predicted class score reductions }
		\State \textbf{for} $s_i$ in $S$ \textbf{do}
		\State\hspace{\algorithmicindent} $c_{new} = {C}_{M}(I \backslash s_i)$ \textcolor{gray}{\% Class after removing $s_i$ from $I$}
		\State\hspace{\algorithmicindent} $p_{c,new} = f_{C_{M,c}}(I \backslash s_i)$ \textcolor{gray}{\% Score after removing $s_i$ from $I$}
		\State\hspace{\algorithmicindent} \textbf{if} $c_{new} \neq c$ \textbf{then}
		\State\hspace{\algorithmicindent}\hspace{\algorithmicindent} $E = E \cup \{s_i\}$
		\State\hspace{\algorithmicindent} \textbf{else}
		\State\hspace{\algorithmicindent}\hspace{\algorithmicindent} $C = C \cup \{s_i\}$
		\State\hspace{\algorithmicindent}\hspace{\algorithmicindent} $P = P \cup (p_c - p_{c,new})$
		\State\hspace{\algorithmicindent} \textbf{end if}
		\State \textbf{end for}
		\State \textbf{while} $E = \emptyset$ \textbf{do}
		\State\hspace{\algorithmicindent} $k =$ argmax($P$)
		\State\hspace{\algorithmicindent} $best = C_k$ \textcolor{gray}{\% Best-first: highest reduction in predicted class score}
		\State\hspace{\algorithmicindent} $best\_set$ = all expansions of $best$ with one segment
		\State\hspace{\algorithmicindent} $C = C \backslash best$ \textcolor{gray}{\% Pruning step}
		\State\hspace{\algorithmicindent} $P = P \backslash p_{k}$ \textcolor{gray}{\% Pruning step}
		\State\hspace{\algorithmicindent} \textbf{for} $C_0$ in $best\_set$ \textbf{do}
		\State\hspace{\algorithmicindent}\hspace{\algorithmicindent} $c_{new} = {C}_{M}(I \backslash C_0)$ \textcolor{gray}{\% Class after removing $C_0$ from $I$}
		\State\hspace{\algorithmicindent}\hspace{\algorithmicindent} $p_{c,new} = f_{C_{M,c}}(I \backslash C_0)$ \textcolor{gray}{\% Score after removing $C_0$ from $I$}
		\State\hspace{\algorithmicindent}\hspace{\algorithmicindent} \textbf{if} $c_{new} \neq c$ \textbf{then}
		\State\hspace{\algorithmicindent}\hspace{\algorithmicindent}\hspace{\algorithmicindent} $E = E\cup C_0$
		\State\hspace{\algorithmicindent}\hspace{\algorithmicindent} \textbf{else}
		\State\hspace{\algorithmicindent}\hspace{\algorithmicindent}\hspace{\algorithmicindent} $C = C \cup C_0$
		\State\hspace{\algorithmicindent}\hspace{\algorithmicindent}\hspace{\algorithmicindent} $P = P \cup (p_c - p_{c,new})$
		\State\hspace{\algorithmicindent}\hspace{\algorithmicindent} \textbf{end if}
		\State\hspace{\algorithmicindent} \textbf{end for}
		\State \textbf{end while}	
		\\
		\State \textbf{\underline{Output:}}
		\State Explanations in $E$
	\end{algorithmic}
\end{algorithm}

\newpage
\section{SEDC-T algorithm} \label{app:sedc_target}
\begin{algorithm}[h]
	\caption{SEDC-T algorithm}
	\label{alg:SEDC_target}
	\begin{algorithmic}[0.6] \scriptsize
		\State \textbf{\underline{Inputs:}}
		\\
		$I$ \textcolor{gray}{\% Image to classify}
		\\
		${C}_{M}: I \rightarrow \{1,2,...,k\}$ \textcolor{gray}{\% Trained classifier with scoring function $f_{C_M}$}
		\\
		$S = \{s_i, i = 1,2,...,l\}$ \textcolor{gray}{\% Segmentation of the image with $l$ segments} 
		\\
		$t$ \textcolor{gray}{\% Target counterfactual class}
		\\
		\State \textbf{\underline{Procedure:}}
		\State $c = {C}_{M}(I)$ \textcolor{gray}{\% Predicted class}
		\State $E = \{\}$ \textcolor{gray}{\% List of explanations}
		\State $C = \{\}$ \textcolor{gray}{\% List of combinations to expand on}
		\State $P = \{\}$ \textcolor{gray}{\% List of differences between target class and predicted class scores}
		\State \textbf{for} $s_i$ in $S$ \textbf{do}
		\State\hspace{\algorithmicindent} $c_{new} = {C}_{M}(I \backslash s_i)$ \textcolor{gray}{\% Class after removing $s_i$ from $I$}
		\State\hspace{\algorithmicindent} $p_{t,new} = f_{C_{M,t}}(I \backslash s_i)$ \textcolor{gray}{\% Target class score after removing $s_i$ from $I$}
		\State\hspace{\algorithmicindent} $p_{c,new} = f_{C_{M,c}}(I \backslash s_i)$ \textcolor{gray}{\% Predicted class score after removing $s_i$ from $I$}
		\State\hspace{\algorithmicindent} \textbf{if} $c_{new} = t$ \textbf{then}
		\State\hspace{\algorithmicindent}\hspace{\algorithmicindent} $E = E \cup \{s_i\}$
		\State\hspace{\algorithmicindent} \textbf{else}
		\State\hspace{\algorithmicindent}\hspace{\algorithmicindent} $C = C \cup \{s_i\}$
		\State\hspace{\algorithmicindent}\hspace{\algorithmicindent} $P = P \cup (p_{t,new} - p_{c,new})$
		\State\hspace{\algorithmicindent} \textbf{end if}
		\State \textbf{end for}
		\State \textbf{while} $E = \emptyset$ \textbf{do}
		\State\hspace{\algorithmicindent} $k =$ argmax($P$)
		\State\hspace{\algorithmicindent} $best = C_k$ \textcolor{gray}{\% Best-first: highest difference between target and predicted class score}
		\State\hspace{\algorithmicindent} $best\_set$ = all expansions of $best$ with one segment
		\State\hspace{\algorithmicindent} $C = C \backslash best$ \textcolor{gray}{\% Pruning step}
		\State\hspace{\algorithmicindent} $P = P \backslash p_{k}$ \textcolor{gray}{\% Pruning step}
		\State\hspace{\algorithmicindent} \textbf{for} $C_0$ in $best\_set$ \textbf{do}
		\State\hspace{\algorithmicindent}\hspace{\algorithmicindent} $c_{new} = {C}_{M}(I \backslash C_0)$ \textcolor{gray}{\% Class after removing $C_0$ from $I$}
		\State\hspace{\algorithmicindent}\hspace{\algorithmicindent} $p_{t,new} = f_{C_{M,t}}(I \backslash C_0)$ \textcolor{gray}{\% Target class score after removing $C_0$ from $I$}
		\State\hspace{\algorithmicindent}\hspace{\algorithmicindent} $p_{c,new} = f_{C_{M,c}}(I \backslash C_0)$ \textcolor{gray}{\% Predicted class score after removing $C_0$ from $I$}
		\State\hspace{\algorithmicindent}\hspace{\algorithmicindent} \textbf{if} $c_{new} = t$ \textbf{then}
		\State\hspace{\algorithmicindent}\hspace{\algorithmicindent}\hspace{\algorithmicindent} $E = E\cup C_0$
		\State\hspace{\algorithmicindent}\hspace{\algorithmicindent} \textbf{else}
		\State\hspace{\algorithmicindent}\hspace{\algorithmicindent}\hspace{\algorithmicindent} $C = C \cup C_0$
		\State\hspace{\algorithmicindent}\hspace{\algorithmicindent}\hspace{\algorithmicindent} $P = P \cup (p_{t,new} - p_{c,new})$
		\State\hspace{\algorithmicindent}\hspace{\algorithmicindent} \textbf{end if}
		\State\hspace{\algorithmicindent} \textbf{end for}
		\State \textbf{end while}	
		\\
		\State \textbf{\underline{Output:}}
		\State Explanations in $E$
	\end{algorithmic}
\end{algorithm}

\newpage
\begin{landscape}
\section{Experiment misclassifications} \label{app:data_misclassifications}
\begin{longtable}{ccccc}
\caption{SEDC-T applied to misclassifications.}\label{tab:misclassifications} \\
			\textbf{Label} 	& \textbf{\# images} & \textbf{Top 5 misclassifications} 	& \textbf{\# misclassified} & \textbf{\# target found}\\
			\hline
			\endfirsthead
			
			\textbf{Label} 	& \textbf{\# images} & \textbf{Top 5 misclassifications} 	& \textbf{\# misclassified} & \textbf{\# target found}\\
			\hline
			\endhead		
			
			acoustic guitar & 2,015								& electric guitar										& 111												& 94\\*
											&											& banjo															& 41 												& 36\\*
											&											& violin														& 21 												& 18\\*
											&											& cello															& 15 												& 13\\*
											&											& stage															& 9 												& 6\\
			\hline
			barrow					& 1,334								& plow															& 20 												& 16 \\*
											&											& park bench												& 17 												& 13\\*
											&											&	tricycle													& 14 												& 12\\*
											&											& horse cart												& 14 												& 13\\*
											&											&	stretcher													& 12 												& 10\\
			\hline
			beach wagon			&	1,360								& minivan														& 99 												& 97\\*
											&											& pickup														& 40 												& 39\\*
											& 										& jeep															& 38 												& 34\\*
											& 										& convertible												& 36 												& 34\\*
											&											& limousine													& 16 												& 13\\
			\hline
			beacon					& 1,806								& breakwater												& 38 												& 37\\*
											& 										& promontory												& 13 												& 13\\*
											& 										& church														& 12 												& 12\\*
											&											& castle														& 11 												& 11\\*
											& 										& bell cote													& 8 												& 8\\
			\hline
			chihuahua				&	1,749								& miniature													& 78 												& 72\\*
											& 										& toy terrier												& 34 												& 33\\*
											& 										& Italian greyhound									& 30 												& 22\\*
											& 										& Boston bull												& 25 												& 21\\*
											& 										& Pomeranian												& 24 												& 18\\
			\hline
			church 					& 1,327 							& castle														& 83 												& 70\\*
											& 										& monastery													& 63 												& 58\\*
											& 										& altar															& 49 												& 46\\*
											& 										& vault															& 44 												& 35\\*
											&											&	bell cote													& 31 												& 29\\
			\hline
			envelope				& 1,023								& wallet														& 28 												& 27\\*
											& 										& carton														& 22 												& 22\\*
											& 										& packet														& 15 												& 15\\*
											& 										& handkerchief											& 8 												& 8\\*
											& 										& binder														& 7 												& 7\\
			\hline
			espresso maker	& 1,126								& coffeepot													& 43 												& 39\\*
											& 										& switch														& 6													& 5\\*
											& 										& Polaroid camera										& 4 												& 3\\*
											& 										& drum															& 3 												& 2\\*
											& 										& printer														& 3 												& 3\\
			\hline
			fire engine			& 1,355								& tow truck													& 36 												& 33\\*
											& 										& garbage truck											& 11 												& 9\\*
											& 										& ambulance													& 6 												& 5\\*
											& 										& thresher													& 6 												& 3\\*
											& 										& harvester													& 5 												& 5\\
			\hline
			meerkat 				& 2,338								& mongoose													& 81 												& 76\\*
											& 										& marmot														& 24 											  & 22\\*
											& 										& Madagascar												& 9 												& 7\\*
											& 										& megalith													& 7 												& 3\\*
											& 										& wallaby														& 6	 												& 6\\
			\hline
			military uniform & 1,430							& bearskin													& 33 												& 20\\*
											&											& assault rifle											& 19 												& 19\\*
											& 										& rifle															& 16 												& 16\\*
											& 										& pickelhaube												& 15 												& 13\\*
										  &											& stretcher													& 11												& 10\\
			\hline
			mouse						& 1,303								& desktop computer									& 65 												& 38\\*
											& 										& desk															& 57 												& 21\\*
											& 										& computer													& 18 												& 17\\*
											& 										& laptop														& 15 												& 11\\*
											& 										& notebook													& 10 												& 6\\
			\hline
			pencil sharpener & 1,268							& switch														& 12 												& 12\\*
											& 										& rubber eraser											& 10 												& 10\\*
											& 										& pencil box												& 9 												& 9\\*
											& 										& Polaroid camera										& 6 												& 6\\*
											& 										& iron															& 6 												& 5\\
			\hline
			Polaroid camera & 1,239 							& reflex camera											& 26 												& 24\\*
											& 										& printer														& 7 												& 6\\*
											&											&	tape player												& 6 												& 5\\*
											& 										& pencil sharpener									& 5 												& 5\\*
											&											& switch														& 4 												& 4\\
			\hline
			revolver				& 1,212								& rifle															& 24 												& 20\\*	
											& 										& assault rifle											& 18 												& 15\\*
											& 										& holster														& 18 												& 15\\*
											& 										& pencil sharpener									& 4 												& 4\\*
											& 										& flute															& 4 												& 3\\
			\hline
			rugby ball			& 1,507								& soccer ball												& 36 												& 29\\*
											& 										& volleyball												& 12 												& 9\\*
											& 										& baseball													& 11 												& 5\\*
											& 										& football helmet										& 9 												& 7\\*
											& 										& cowboy hat												& 6 												& 5\\
			\hline
			soccer ball			& 1,344								& rugby ball												& 11 												& 11\\*
											& 										& volleyball												& 9 												& 7\\*
											& 										& ballplayer												& 7 												& 4\\*
											& 										& unicycle													& 7 												& 4\\*
											& 										& croquet ball											& 6												  & 6\\
			\hline
			tiger						& 2,085								& tiger cat													& 30 												& 30\\*
											& 										& snow leopard											& 14 												& 12\\*
											& 										& lynx															& 11 												& 8\\*
											& 										& cougar 														& 8 												& 8\\*
											& 										& zebra															& 8 												& 8\\
			\hline
			toaster 				& 1,357								& microwave													& 17 												& 14\\*
											& 										& pencil sharpener									& 16 												& 15\\*
											&  										& rotisserie												& 7 												& 7\\*
											& 										& switch														& 6 												& 5\\*
											& 										& printer														& 6 												& 5\\
			\hline
			warplane				& 1,097								& aircraft carrier									& 33 												& 30\\*
											& 										& airliner													& 22 												& 22\\*
											& 										& space shuttle											& 16 												& 16\\*
											& 										& wing															& 12 												& 12\\*
											& 										& missile														& 7	 												& 5\\
			\hline	
\end{longtable}

\end{landscape}

\end{document}